\documentclass{article}
\usepackage[utf8]{inputenc}
\usepackage[english]{babel} 
\usepackage{comment}
\usepackage{graphicx} 
\usepackage[titletoc,title]{appendix} 
\usepackage{caption} 
\usepackage{natbib}
\usepackage[nottoc]{tocbibind} 
\usepackage{perpage} 
\usepackage{color,soul} 
\usepackage{longtable}
\usepackage{url}
\usepackage[caption=false]{subfig}

\def\eg{e.g., } 
\graphicspath{{images/}}

\title{Frugal Machine Learning}

\author{
    Mikhail Evchenko\\
    \texttt{mikhailevchenko@gmail.com} 
    \and
    Joaquin Vanschoren\\
    \texttt{j.vanschoren@tue.nl}
    \and
    Holger H. Hoos\\
    \texttt{hh@liacs.nl}
    \and
    Marc Schoenauer\\
    \texttt{marc.schoenauer@inria.fr}
    \and
    Mich\`{e}le Sebag\\
    \texttt{michele.sebag@lri.fr} 
 } 

\date{} 
\newcommand{\hide}[1]{}

\begin{document}
\maketitle

\begin{abstract} 
Machine learning, already at the core of increasingly many IT systems and applications, is set to become even more ubiquitous with the rapid rise of wearable devices
and the Internet of Things. 
In most machine learning applications, the main focus is on the quality of the results achieved (e.g., prediction accuracy), and hence vast amounts of data are being collected, requiring significant computational resources to build models. 
In many scenarios however, it is infeasible or impractical to set up large centralized data repositories. In personal health, for instance, privacy issues may inhibit the sharing of detailed personal data. In such cases, machine learning should ideally be performed on wearable devices themselves, which raises major computational limitations such as the battery capacity of smartwatches. This paper thus investigates {\em frugal learning}, aimed to build the most accurate possible models using the least amount of resources. A wide range of learning algorithms is examined through a `frugal lens', analysing their accuracy/runtime performance on a wide range of data sets. The most promising algorithms are thereafter assessed in a real-world scenario by implementing them in a smartwatch and letting them learn activity recognition models on the watch itself.
\end{abstract}

\section{Introduction}
Machine learning (ML) methods are rapidly gaining traction and importance in virtually all applications of information technology. The success and pervasive use of machine learning techniques is fuelled by a combination of increasing availability of data and readily accessible, cheap and powerful computational resources. While applications to very large amounts of data remain computationally challenging, computational resources are no longer a limiting factor in the general case\footnote{An exception might be that of Deep Learning \citep{Bengio}.}. 

Given the number and diversity of the ML algorithms, system developers and users face the long known issue of {\em algorithm selection}, aimed at selecting the algorithm with best expected performance on the considered use case. Performance indicators are most usually related to the prediction quality (expected fraction or cost of errors), regardless of running time. 

However, there are increasingly many use cases subject to different performance measures. 
For example, when running code on portable devices, particularly mobile phones,
conserving battery charge is still an important objective.
This holds to an even larger extent for wearable devices, such as smartwatches, which additionally are much more limited in computational power, and for most of the computing technology involved in the Internet of Things (IoT).
For these application contexts, in addition to limitations on battery capacity and computational power, data transmission is typically limited or costly.
While it can be expected that all of these limitations will become less stringent as technology continues to progress, they will remain, at least in part, relevant in the near and perhaps not-so-near future. 

Another consideration that has the potential to impose fundamental constraints on the use of ML algorithms is privacy, related to \eg  personal health, activities or habits. Not everyone feels comfortable to share
fine-grained data of this nature with a service provider, in order to get better advices or recommendations. In many cases, obtaining slightly less accurate feedback based on substantially less
sensitive personal data can be appealing. Along this line, a trade-off is 
offered by processing private data on-board (using a portable or wearable device such as a smartwatch) and offering ML services without transmitting sensitive data at all, thereby addressing privacy concerns and reducing the need for data transmission. Similar considerations apply to smart home applications, embedded sensing and control systems, as well as in the context of many IoT scenarios.

These considerations motivate the concept of \emph{frugal machine learning},
which emphasises the cost associated with the use of data and computational resources.
Frugality, i.e., the idea of working with limited resources, comes in different flavours:
\begin{itemize}
    \item Input frugality emphasises the cost associated with the data, specifically with the acquisition of the training data, the exploitation of the descriptive features, or both. Frugal inputs may involve fewer training data or fewer features than required for the best prediction quality achievable in a non-frugal setting.
    Input frugality can be motivated by resource constraints and by privacy constraints. 
    \item Learning process frugality emphasises the cost associated with the learning process, specifically the computational and memory resources. Frugal learning might produce a model with lower prediction quality than achievable in a non-frugal setting, but do so much more efficiently. Learning process frugality is primarily motivated by resource constraints, including limited computational power and limited battery capacity.
    \item Model frugality emphasises the cost associated with storing or using a machine learning model, such as a classifier or regression model. For supervised learning, frugal models may require less memory and produce predictions with less computational effort than required for optimal prediction quality.
    Model frugality is primarily motivated by resource constraints such as limited memory or limited processing capabilities.
\end{itemize}

This paper is motivated by the new ML settings induced by on-board wearable devices such as smartwatches, and similarly limited devices, facing severe restrictions in terms of i) acquisition and transmission of data, due to cost or privacy issues; ii) learning and decision making, due to energy issues. 

We focus on frugal supervised learning, and specifically classification. Our goal is to empirically investigate the trade-off between predictive accuracy and the computational cost involved in learning models and using them, for a wide range of ML algorithms and datasets. The motivating application is that of a smartwatch used as a stand-alone device for activity recognition, using supervised classification to recognise various types of activities, such as running, walking, bicycling, driving, riding a bus, weightlifting, resting or sleeping, from biometric and other sensor input.

The remainder of this article is structured as follows.
Section~2 provides a formal background and briefly discusses related work. 
In Section~3, we motivate and introduce a quantitative measure for frugality, which we then use to study a broad range of classification algorithms and benchmarks from the well-known OpenML platform for reproducible machine learning research \citep{vanschoren2014openml}.
Section~4 is devoted to the experimental setting. Section~5 reports on the results of our empirical study. A striking (though not unexpected) result is how strongly algorithm rankings depend on the desired trade-off between predictive accuracy, computational learning cost, and decision making cost. 
The paper concludes with the lessons learned regarding the trade-off between 
predictive performance and frugality, and discusses our perspectives for further work.

\hide{earlier outline:
[HH to start fleshing out some text]
motivation:
- wearable devices (-> restricted resources, especially power -> running time, 
but also memory)
- health data -> privacy
- smart home -> privacy [relationship wearables: e.g., in-home local tracking using
bluetooth le = ibeacon]
- activity recognition for elders (anomaly detection)
- in-store / in-area advertising / shopping assistance -> privacy

- tie to concrete example on actual wearable?

what is frugal ml, and why is it important?
- cost of features (human / machine effort of determining them)
- cost of running the model (running time -> power, memory)
- cost of learning the model (running time -> power, memory) 
- privacy: features (i.e., input to the model),
meta-features (i.e., input to a meta-learning procedure that helps to 
shortcut learning)
model (may not want to share the model with anyone)
-> might have to do learning on device
- important because of on-device / privacy issues (see motivation)
- conceptual:
- frugal inputs (features: time, expertise, privacy)
- frugal models (outcome)
- frugal learning (process)
- frugal meta-learning (meta-features, meta-learning process)

(briefly position against most closely relatd work, point to sec 2 for detailed discussion)

(briefly outline main technical contributions)

here: restrict to classification settings (to keep things manageable)
}

\section{Related work}
\label{sec:related}

Frugality has been considered in Machine Learning with regard to four different goals.
The first three goals consider the memory and/or computation and/or oracle resources required to achieve learning and build a hypothesis. The fourth goal considers the computational resources required to make a decision based on a learned hypothesis. 

To our best knowledge, frugal learning-oriented research has been devoted to supervised learning, the best and longest established ML task; supervised ML will be the only learning task considered in the remainder of this article.

\def\x{\mbox{\bf x}}
\def\RR{{\rm I\hspace{-0.50ex}R}}
\subsection{Notations}
\label{subsec:notations}
Let $\cal E$ denote the training set, made of $n$ pairs $(\x_i,y_i)$, with instance $\x_i$ in the instance space $\cal X$ (in the vast majority of cases, a propositional space e.g. $\RR^d$; else a relational space, e.g. a logic program), and $y_i$ in the label space $\cal Y$, either binary $\{-1,1\}$ or multi-class $\{0,1, \ldots K \}$ or real-valued. The characteristics of instances (the coordinates, in the $\RR^d$ case) are referred to as {\em features}.

Let $\cal H$ denote the hypothesis space, mapping $\cal X$ onto $\cal Y$.

The learned hypothesis $h$ is most usually built by solving an optimization problem, which involves a data fitting term ${\cal F}(h,{\cal E})$, measuring how closely $h(\x_i)$ fits $y_i$
for $1 \le i \le n$, and a regularization or penalization term ${\cal R}(h)$, meant to address overfitting\footnote{Overfitting characterizes the discrepancy between $h$ performance on the training set $\cal E$ used to build $h$, and its performance on further data, say another independent training set ${\cal E}'$, which has not been used to build $h$. The learning goal is to optimize  the expectation of ${\cal F}(h,{\cal E}')$, referred to as the generalization error of $h$.
}, and/or enforcing $h$ compliance with prior knowledge or additional requirements. 

\[ h = \arg\min_{h \in {\cal H}} \{ {\cal F}(h,{\cal E}) + {\cal R}(h) \} \]

As claimed by \cite{BottouB07}, all dimensions of frugality, ranging from example and model selection to learning computational cost are relevant to the quality of the learned models. 

\subsection{Frugality through compression or approximation} 
\label{subsec:decision_making}
Early works related to frugal learning achieved the compression of 
the training set $\cal E$ itself. Formally, the goal was to map the training set, viewed as a (partial) decision table, into a resource-bounded device. 
This approach is currently blossoming in the field of binary decision diagrams, pioneered by \cite{Minato}. Note that binary decision diagrams can be directly implemented into Field Programmable Gate Arrays (FPGAs) \citep{FPGA}. 

More remotely related is the work on approximate knowledge bases \citep{Marquis,Darwiche}. The hypothesis is formed of a large or complex knowledge base $KB$, \eg\ with severely limited real-time exploitation. The approach aims at building a (significantly smaller) lower and an upper bounded KB, respectively denoted $KB_L$ and $KB_U$, used to approximate the $KB$ decision, respectively holding $\x$ as true if $KB_L(\x)$ is true, false if $KB_U(\x)$ is false, and unknown otherwise. 

\subsection{When frugality helps learning}

An essential concern for Machine Learning is to avoid overfitting, with hypothesis frugality as a consequence: the more frugal the hypothesis, the less likely it will overfit the data. In early days, an empirical strategy was to detect and remove the hypothesis components which were the less active, e.g. cancelling out the weights with lowest absolute value in a neural net, a heuristic referred to as {\em Optimal Brain Damage} \citep{OBD}.
Later on, with the advent of the structural risk minimization \citep{Vapnik92,Vapnik95}, a regularization term was added to the data fitting objective to reduce the variance, and henceforth
the richness, of the learned hypothesis. 

\begin{figure}
\centering 
\includegraphics[scale=0.38]{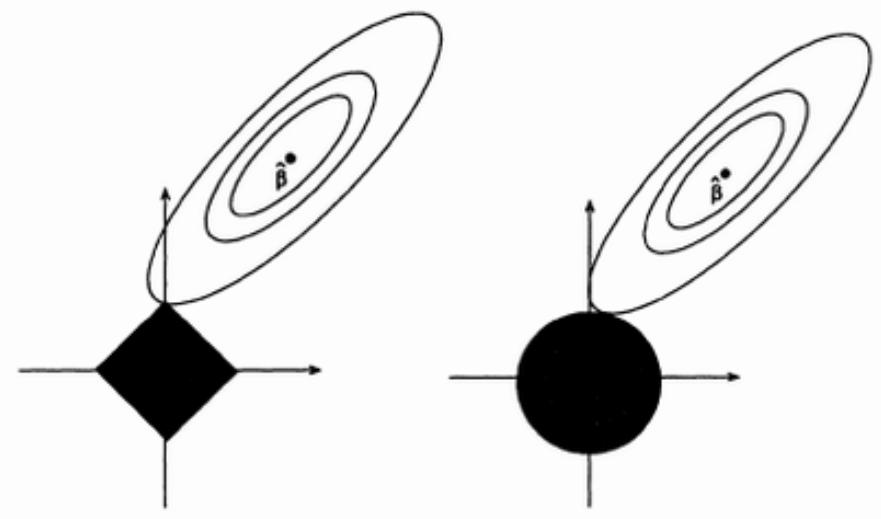}
 \caption{Isolines of the data fitting term: feature selection  with $L_1$ (left) and $L_2$ (right) regularization}
 \label{fig:L1-L2}
\end{figure}

Some regularization terms directly support frugality through feature selection, reducing the number of features involved in the hypothesis.
For instance, regularization terms $R(h)$ based on the $L_1$ norm of $h$ (as opposed to $L_2$ norm) typically contribute to minimizing the number of features involved in hypothesis $h$. This can be graphically understood as shown in Fig. \ref{fig:L1-L2}: the isolines of the data fitting term ${\cal F}(h,{\cal E})$ are more likely to cut the $L_1$ ball at a corner (which can be expressed in fewer features), than for an $L_2$ ball. 

A huge literature is devoted to feature selection (FS). The main three approaches involve: \textit{scoring or filtering approaches}, which independently rank features using some score function and select the top-ranked ones; \textit{wrapper approaches}, which tackle the whole combinatorial optimization problem of finding the subset of features which optimize (an estimate of) the generalization error; \textit{embedded approaches}, which combine learning and feature selection through prior regularization or posterior pruning \citep{Tibshirani,RFE}. 

Another ML field related to frugality is that of online learning, where examples are considered one at a time, and used to update the current hypothesis (see \cite{Online14} and references therein). The key issue is related to the possible drift of the sample distribution, and the subsequent adaptation of the hypothesis. The trade-off here concerns the number of examples used by the learner: distribution changes should be detected as early as possibly (and outdated examples discarded) while the learned hypothesis must be robust against example noise. Moreover, several online algorithms for high-speed data streams only use a subsample of the available examples in order to be trained faster \citep{Domingos:2000:MHD:347090.347107}.

\subsection{Learning with frugal oracle resources} 
\label{subsec:learning_with_frugal} 
ML performance classically improves, everything else being equal, as the amount of training examples increases. The main request is on the number of {\em labelled} examples; in many applications, unlabelled examples can be gathered with virtually no limits, e.g. on the Web or by observing the phenomenon under study. The shortage is on {\em labelled} examples, either because it requires expertise, or because it requires additional expensive procedures (e.g., medical tests, destructive experiments). 

Active learning is concerned with the selection of the most informative samples. These samples are labelled by the oracle or teacher, and the learning process goes on in an online fashion. Quite a few criteria have been designed for active learning; typically the samples that are the most uncertain according to the current hypotheses bring more information. In an ideal case, selecting the most informative examples might yield and exponential increase of the learning speed (see \cite{Dasgupta11,ShalevShwartz13} and references therein); in practice however, the most informative examples might also be the most difficult ones for the oracle to label, slowing down the process. 

\subsection{Decision making and frugality}

Some ML attempts toward frugal decision making consider a two-step process. In a first step, a possibly computationally expensive model is learned; in a second step, this model is simplified in diverse manners while preserving its predictive accuracy to the best possible extent. 

In \citet{RFE} for instance, the features involved in a linear classifier with low absolute weight value are considered to be poorly relevant features (or redundant with other ones), and they are removed, possibly along an iterative process. \cite{Busa-FeketeBK12} consider the (usually large) set of hypotheses learned along a boosting process. Traditionally, these hypotheses vote to deliver the final decision. It is possible, however, to prune the set of hypotheses in an example-dependent manner, converting the ensemble into a Directed Acyclic Graph. 

A most striking example is due to \cite{Caruana}, aimed at compressing deep neural nets \citep{Bengio}. 
Let us consider a (large) ensemble of (very large and deep) neural networks, referred to as the teacher, and learned from some training set $\cal E$. This ensemble is used to label a much larger set of samples ${\cal E}'$. \cite{Caruana} show that, using this new training set (made of the samples in ${\cal E}'$ and the associated teacher label), a {\em shallow} neural net with same performances as the teacher can be trained, with orders of magnitude gains on the number of weights (though the process cannot be considered frugal).

\section{How to empirically assess frugality}
\label{sec:how_to_empirically}

As was shown in Section \ref{sec:related}, several methods and strategies can be considered to perform machine learning on a resource limited device. To find out exactly how \textit{frugal} each of these are, and which ones to select for applications, we need to empirically assess their performance. This can be done by running the machine learning techniques on the device, and measuring their predictive performance as well as their usage of resources, i.e. CPU cycles or runtime, RAM usage, or battery consumption. 

Analyzing all of these factors independently is very useful, but can also be quite cumbersome. One can find that method A is very accurate but slow, and method B is less accurate but fast: which method is most frugal? Moreover, these factors may actually interact with each other. For instance, on smartwatches, sensors often produce data on a best-effort basis depending on the available CPU cycles and battery charge. If either of these is low, they can produce less frequent data, ultimately harming the predictive performance of machine learning applications that depend on that sensor information. Hence, an algorithm that is very accurate (but CPU-hungry) on a smartphone may be significantly less accurate when running on a smartwatch with more limited resources.

Hence, it is useful to define a novel, multi-objective evaluation measure, i.e. \textit{frugality}, that combines both predictive performance and resource consumption, and allows us to easily compare different techniques. Ideally, it also includes a parameter that allows us to adjust the need for frugality by the device, e.g., depending on whether it is a powerful server, a smartphone, or small chips in wearable devices.

To make this more concrete, consider the problem of selecting a classification algorithm for use on a wearable device. One could consider relatively simple algorithms, such as 
Decision Trees \citep{quinlan2014c4}, which can be constructed fast and without substantial memory allocation. On the other hand, Decision Trees are often outperformed by more sophisticated but slower algorithms such as Support Vector Machines \citep{vladimir1995nature} or Neural Networks \citep{Schmidhuber201585}. Consequently, there exists a clear trade-off between predictive performance (e.g., accuracy) and usage of resources (CPU and RAM).

\subsection{Existing multi-objective measures}
There are many different ways to combine multiple measures into a single one. In the context of algorithm selection, \citet{brazdil2003ranking} have proposed the Adjusted Ratio of Ratio's (ARR). This metric, however, has the drawback that the constructed function is not monotonic, and thus hard to interpret \citep{abdulrahman2014measures}. To address this problem, \citet{abdulrahman2014measures} proposed the A3R measure, defined as follows:

\begin{equation}
    \mathit{A3R}_{a_\mathit{ref},a_j}^{d_i} = 
      \frac{ \frac{\mathit{SR}_{a_j}^{d_i}}
                    {\mathit{SR}_{a_\mathit{ref}}^{d_i}}} 
            { \sqrt[N]{T_{a_j}^{d_i} / T_{a_\mathit{ref}}^{d_i}}}
\end{equation}

Here, $\mathit{SR}_{a_j}^{d_i}$ and $\mathit{SR}_{a_\mathit{ref}}^{d_i}$ represent the \emph{success rates} (e.g. predictive accuracy) of algorithms $a_j$ and $a_\mathit{ref}$ on data set $d_i$, where $a_\mathit{ref}$ represents a given \emph{reference algorithm} for pairwise comparison. Similarly, $T_{a_j}^{d_i}$ and $T_{a_\mathit{ref}}^{d_i}$ represent the run times of the algorithms, in seconds. To trade off the importance of time, A3R includes the $N^{th}$ root parameter: the higher the value for $N$, the smaller the influence of runtime.

A simplified, non-pairwise version of A3R introduced by~\citet{Rijn2015} assumes that both the success rate of the reference algorithm $\mathit{SR}_{a_\mathit{ref}}^{d_i}$ and the corresponding time $T_{a_\mathit{ref}}^{d_i}$ have a fixed value, set to 1. This version, called $A3R'$, is defined as follows: 

\begin{equation}
    \mathit{A3R'}_{a_j}^{d_i} = 
      \frac{ \mathit{SR}_{a_j}^{d_i}
                    } 
            { \sqrt[N]{T_{a_j}^{d_i}}}
\end{equation}
z
This is a useful measure of frugality, but it can be hard to choose a sensible value for N given that it ranges between 1 and infinity. Moreover, it reaches a value close to infinity for very small runtimes, which is not very intuitive.

\subsection{Frugality score}
In our definition of frugality, we would like a score function that is bounded within a fixed range, e.g. [-1,1]. We do still want a trade-off parameter between predictive performance and runtime, but ideally this value can be adjusted between 0 and 1. These constraints led us to the following definition: 

\begin{equation} \label{eq:frugal_score} 
Frug_{a_j}^{d_i} = P_{a_j}^{d_i} - \frac{w}{1 + \frac{1}{R_{a_j}^{d_i}}}
\end{equation}

where $P_{a_j}^{d_i}$ is a measure of predictive performance of algorithm $a_j$ on data set $d_i$ that needs to be maximized, w is tradeoff coefficient that defines the importance (or \textit{weight}) of frugality (i.e., how scarce resources are), and $R_{a_j}^{d_i}$ is the resource consumption of algorithm $a_j$ on data set $d_i$. 

In this paper, we will mainly use the Area under the ROC curve \citep{bradley1997use}, or AUC, as the measure for $P_{a_j}^{d_i}$, because it is fairly robust against imbalanced classes. For multiclass problems, we use a multiclass AUC \citep{hand2001simple} that uses a one-versus-all approach for every class value. For $R_{a_j}^{d_i}$, we will mainly use CPU time (in milliseconds, and non-zero), or more precisely the sum of the training time and prediction time, since both have to be performed on the wearable device, and CPU usage affects battery life most. Hence, we will be using $Frug_{a_j}^{d_i} = AUC - \frac{w}{1 + \frac{1}{T_{train}+T_{test}}}$ in our experiments.

Note however, that these can easily be substituted by other measures depending on the application. For instance, if memory consumption is an issue, one could use a measure similar to Ram-Hours \citep{Bifet:2013c}, which is simply the product of the RAM usage (in GB) and the CPU time (in hours).

In this formula, $w$ defines the demand for frugality. The range for $w$ can vary from 0 to infinity, but in practice we will use values between 0 and 1. If $w$ is high, e.g. $w = 1$, our frugality score will favor algorithms that run fast with reasonable accuracy, useful for wearable devices. Values higher than 1 indicate that predictive performance is less important. When $w$ is low, e.g. 0.1, then priority is given to algorithms with better performance, even if they are slower. This is useful for smartphones or more powerful devices. If the value of $w$ is set to 0, then algorithms will be ranked only by AUC without taking into account the required training and testing time, as may be the case in cloud computing.
 
The derivation of $\frac{w}{1 + \frac{1}{R}}$ is as follows: we want to scale resource consumption between 0 and 1, and hence transform it using sigmoid function scaled to that range: $\frac{1}{1 + \exp(-R)}$. However, since runtimes often grow exponentially, we take a logarithm, yielding $\frac{1}{1 + \exp(-\log(R))}$. This is equivalent to the frugality definition above. Using AUC and a value for w between 0 and 1, the frugality score will be a value between -1 and 1.

\begin{figure}[t] 
\centering 
\includegraphics[scale=0.38]{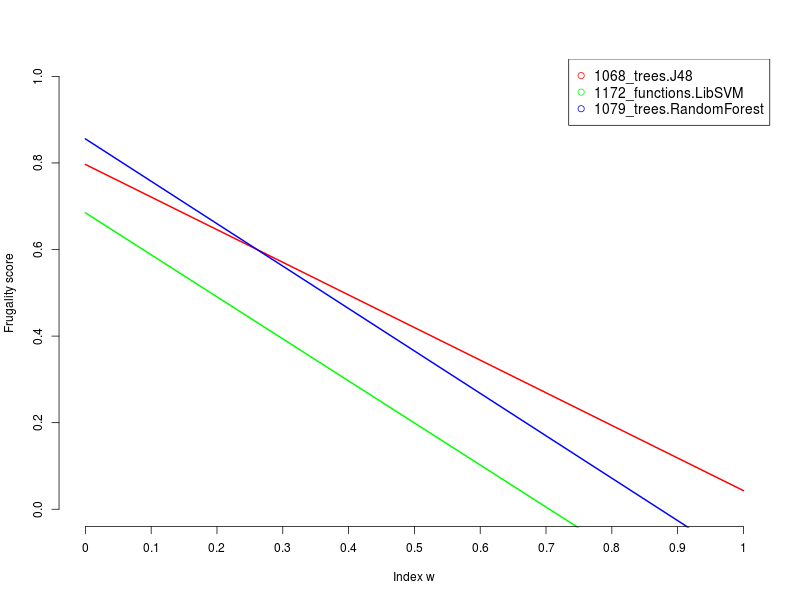}
\caption{Illustrative frugality curves for Decision Trees (J48), Random Forest, and LibSVM (SVM with RBF kernel).} 
\label{fig:frugal_score_example}
\end{figure}

This definition of frugality can be used to create \textit{frugality curves} to analyse the frugality of different algorithms in function of w, as shown in Figure \ref{fig:frugal_score_example}. Three algorithms were selected to show how the frugality score changes as we put more emphasis on resource constraints. Note that these results are calculated based on the averaged performance and runtime for each algorithm over a wide range of data sets. In other words, these lines are the average of all lines obtained for the same algorithm over all data sets.

Frugality curves also show how algorithm rankings change as frugality becomes more important: the values for $w=0$ represent the ranking of all algorithms based on AUC alone. As we increase the value for $w$, we can see how this ranking changes. Hence, when computational resources are not an issue you would consider algorithms following the ranking for $w=0$, while on a smartwatch you would consider their ranking for a higher $w$ value. 

Considering the algorithms in Fig. \ref{fig:frugal_score_example}, when w is equal to zero, Random Forests perform best overall, followed by Decision trees and LibSVM (using an RBF kernel).\footnote{Note however, that we are using default parameter settings here. In future work, we will optimize algorithms on each data set individually.} When frugality becomes more important ($w$ is higher), the ranking of algorithms changes. Especially the curve crossings are useful, as they indicate the point where you would replace one algorithm with another as resources become more scarce. Indeed, while Random Forests initially performed best, it is overtaken by Decision Trees around $W=3$. LibSVM never overtakes the other algorithms, and drops below zero around $w=0.75$.

A much larger set of algorithms will be analysed in Section \ref{subsec:perf_evaluat}.  

\section{Experimental setup}
\label{sec:exper_set}

To be able to perform a detailed analysis of many learning algorithms in terms of frugality on a wide range of data sets, we downloaded around 53,000 evaluations (both AUC and runtime) of 103 algorithms on 517 data set from OpenML\footnote{http://www.openml.org}, a collaborative platform for reproducible machine learning research \citep{vanschoren2014openml}. The size of these data sets varies from 10 to 98528 observations. The algorithms all originate from the WEKA \citep{hall2009weka} library. A full list of algorithms and used parameter settings is available in Appendix \ref{appendix:algorithms_studied}. Note that a few algorithms don't have evaluations on all datasets, typically due to execution errors or timeouts. If an algorithm has more than 10 missing results, it was deleted from further processing. The names of removed algorithms and the amount of missing values per algorithm are shown in Appendix \ref{appendix:missing_values}. 

The remaining missing values were imputed using the Iterative SVD method \citep{fuentes2006using} from the R package SpatioTemporal\footnote{https://cran.r-project.org/web/packages/SpatioTemporal/index.html}. This method computes missing values by generating a singular value decomposition (SVD) of the original matrix and multiplying the resulting matrices again using the first n components of SVD. The reason of using this method instead of taking the mean performance for an algorithm over all data sets is because every data set is different, and hence AUC and time for each data set are not normally distributed. The Iterative SVD produces more reliable values, since it takes the values of the other algorithms over these data sets into account.  

In terms of computational resources, most algorithms were run on 8-core machines operating at 2GHz, 12 GB memory, and Fedora 14 64-bit as operating system. OpenML also keeps system benchmarks for each system using the SciMark benchmarking tool\footnote{SciMark: \url{http://math.nist.gov/scimark2/}} for future reference. In addition, we ported a selection of WEKA algorithms to run on an Android-based smartwatch. We used an LG URBANE smartwatch with Qualcomm Snapdragon 400 CPU, 512 MB RAM and 410 mAh li-Ion battery.

For the exploratory analysis R version 3.2.2 was used. The R ecosystem provides access to numerous packages that allow making analysis, validating results and visualizing them. The `OpenML' package was used for downloading prior machine learning experiments and uploading new results.


\section{Empirical Analysis}
\label{sec:exper_res}

We used different techniques to study machine learning algorithm performance through a ‘frugal lens’. The goal is to understand the trade-offs between predictive performance and resource usage for a wide range of algorithms, and select the algorithms that are most useful for a given scenario. Ultimately, our goal is to identify a very small number of algorithms that are subsequently implemented on wearable devices for further analysis and use.

In this section, we will perform various experiments. First, since the frugality of an algorithm may depend on properties of the data set, e.g. an algorithm may be equally accurate but much slower on high-dimensional data, it behooves us to assess the frugality of algorithms on different types of data sets separately. In Section~\ref{subsec:an_data_set}, we therefore first cluster the data sets based on the performance of the algorithms trained on them. Second, we will do a Pareto Front analysis of our machine learning algorithms in terms of predictive performance and the CPU time required for running them in Section~\ref{subsec:pareto_front}, to determine which algorithms are most interesting to study further. 
Next, in Section~\ref{subsec:hierar_clust_algorithms}, we generate hierarchical clusterings of data sets to visualize the performance of these algorithms on all data sets for different levels of frugality. Finally, in Section~\ref{subsec:perf_evaluat}, we present frugality plots to analyse a subset of representative algorithms and data sets.

\subsection{Analysis of data sets}
\label{subsec:an_data_set}

Analysing a broad variety of data sets is important for obtaining general insight into the performance of algorithms. If one wants to identify the most appropriate algorithms for different frugality levels, it is necessary to build a study that comprises different data collected for various tasks. OpenML, a platform for reproducible machine learning research, provides the opportunity to get this data with minimal effort. One can create collections of data sets based on different data set properties. In this study we focused on obtaining a very diverse set of classification data sets in terms of number of examples, dimensions and classes. The complete list of data sets, algorithms, and experiments can be found online\footnote{\url{http://www.openml.org/s/1}}. All together, 517 data sets were selected for this study. For each data set, OpenML provides around 100 data characteristics (meta-features). In this paper, we will use a subset of them, presented in Table \ref{tab:attributes}. 

\begin{table}[!htb] 
\centering
\caption{Attributes of data sets and their descriptions}
\label{tab:attributes} 
\begin{tabular}{|p{3cm}|p{8cm}|}
\hline
\textbf{Name of parameter}           & \textbf{Description} 
\\ \hline
NumAttributes               & The number of attributes in a data set. 
\\ \hline
ClassEntropy                & Entropy of the class attribute. This parameter defines the amount of information required to identify the class of an instance. If entropy is low, then the data set has a skewed class distribution. 
\\ \hline
MaxNominalAtt-DistinctValues & The maximum number of distinct values found in the categorical attributes. 
\\ \hline 
MajorityClassSize           & The number of instances (examples) in a data set labeled with the majority (the most frequent) class. 
\\ \hline   
DecisionStumpAUC            & The AUC performance of a 1-level decision tree (Decision Stump) trained on the data set. Also known as a landmarker or probing feature.
\\ \hline
\end{tabular} 
\end{table} 

\subsubsection{Separating data sets in clusters} 
\label{subsubsec:sep_dat_set} 

First, we will cluster the data sets based on the performance of the algorithms trained on them, motivated by the question whether the algorithms behave similarly on certain groups of data sets. Once we know this, we can study the frugality of algorithms on specific types (clusters) of data sets.

To do this, a first question that we want to answer is whether there exists some level of structure in the collection of data sets, or whether they are randomly distributed. The Hopkins statistic \citep{banerjee2004validating} can be used to analyse how much structure there is in our meta-data (the space of algorithm performance data). This statistic generates randomly and uniformly distributed points and measures how well these points fit into the existing data. When distances between randomly generated and existing points are approximately the same, then the value of the index lies around 0.5. The further the value deviates from 0.5, the more structure exists in the data. On our data, we measure a Hopkins statistic of 0.032, which strongly suggests that the meta-data for our data sets is not randomly distributed.

\begin{figure}[!htb] 
\centering 
\includegraphics[scale=0.41]{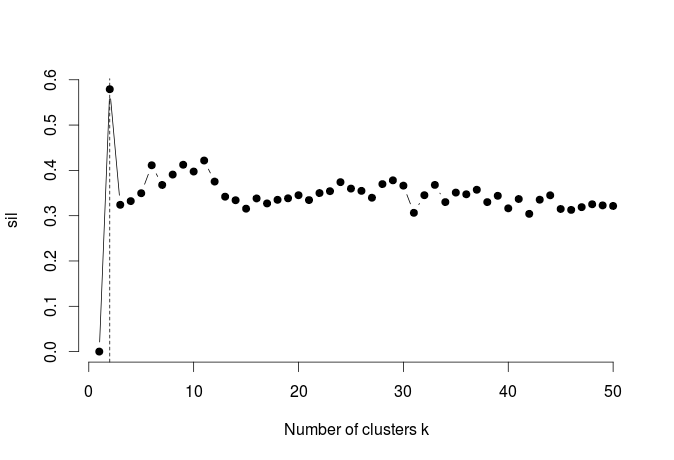}
\caption{Number of clusters based on Silhouette method.} 
\label{fig:silhouette}
\end{figure} 

Having determined that our meta-data has structure, and therefore has clusters, 
the next step is to identify the expected number of clusters. This can be done with help of the Silhouette method \citep{rousseeuw1987silhouettes}, shown in Formula~\ref{eq:silhouette}, which assesses quantitatively how well a given set of data points can be separated into a specific number of clusters.   

Value $a(i)$ in Formula~\ref{eq:silhouette} is the average dissimilarity of point $i$ to all other points within the same cluster, and $b(i)$ is the average dissimilarity to the nearest cluster that point $i$ does not belong to. 

\begin{equation} \label{eq:silhouette} 
s(i) = \frac{b(i) - a(i)}{\max \{a(i), b(i)\}}  
\end{equation} 

Figure~\ref{fig:silhouette} presents the Silhouette values for different numbers of clusters, showing that the highest value of $s(i)$ was achieved when the number of clusters was equal to 2. Therefore, we will focus on clustering the data sets using kMeans clustering \citep{hartigan1979algorithm} with 2 clusters. We also used hierarchical clustering, which will be discussed in Section~\ref{subsec:hierar_clust_algorithms}.

\subsubsection{Visualizing clusters}
\label{subsubsec:visual_clust} 

Based on the previous analysis, we can now build and visualize a clustering of our data sets using kMeans with $k=2$. Popular algorithms for visualizing high-dimensional data are Principal Component Analysis \citep{jolliffe2002principal} (PCA) and t-SNE \citep{van2008visualizing}. These algorithms are based on different approaches for constructing a visualization and provide different perspectives on the clustering.

\begin{figure}[p] 
\centering 
\includegraphics[scale=0.42]{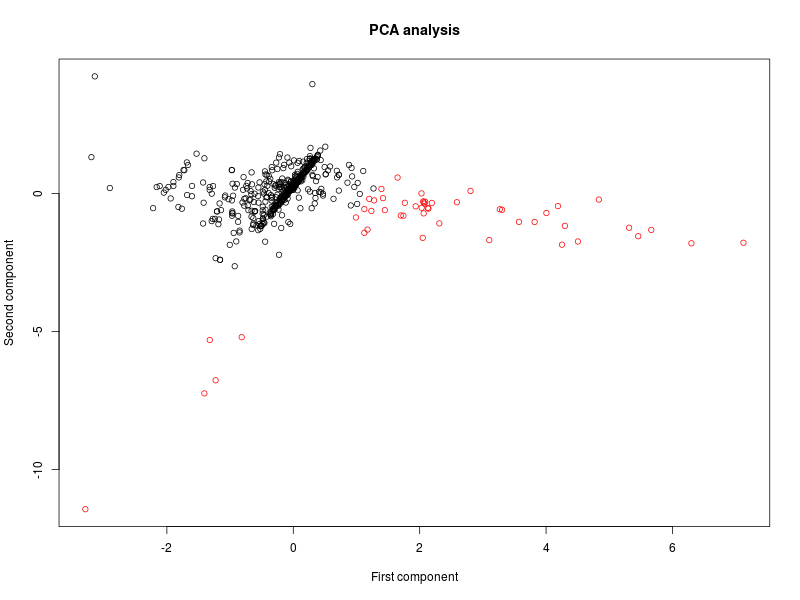}
\caption{Visualization based on first two principal components from PCA.} 
\label{fig:pca}
\includegraphics[scale=0.42]{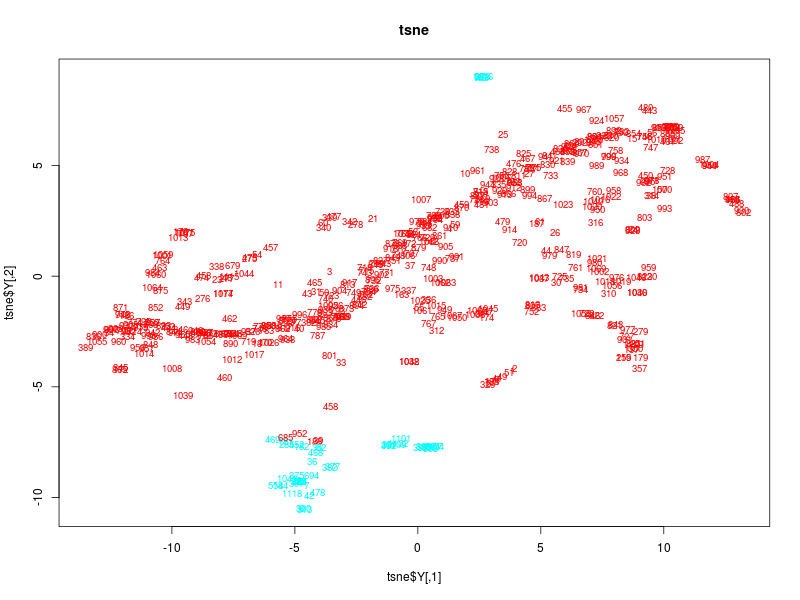}
\caption{The result of t-SNE visualization.}  
\label{fig:tsne}
\end{figure} 

Figure~\ref{fig:pca} visualizes the data set clusters using the first two principal components of PCA. It shows a larger, dense cluster (black dots), and a second, more sparse cluster (red dots). Both clusters can be cleanly separated. 

Figure~\ref{fig:tsne} visualizes the same clusters using t-SNE. We again find that the large cluster (red) and the smaller cluster (blue) can be cleanly separated. Moreover, here we show the OpenML IDs of the data sets. For instance, details on data set 33 can be found at \url{www.openml.org/d/33}. 

The results from Figure~\ref{fig:pca} and Figure~\ref{fig:tsne} demonstrate that our data sets can indeed be cleanly separated into two clusters. Note, however, that results obtained with t-SNE can be prone to noise. Therefore, we verify the robustness of the t-SNE results by adding a noise feature to our meta-data.

\begin{figure}[!htb] 
\centering
\includegraphics[scale=0.29]{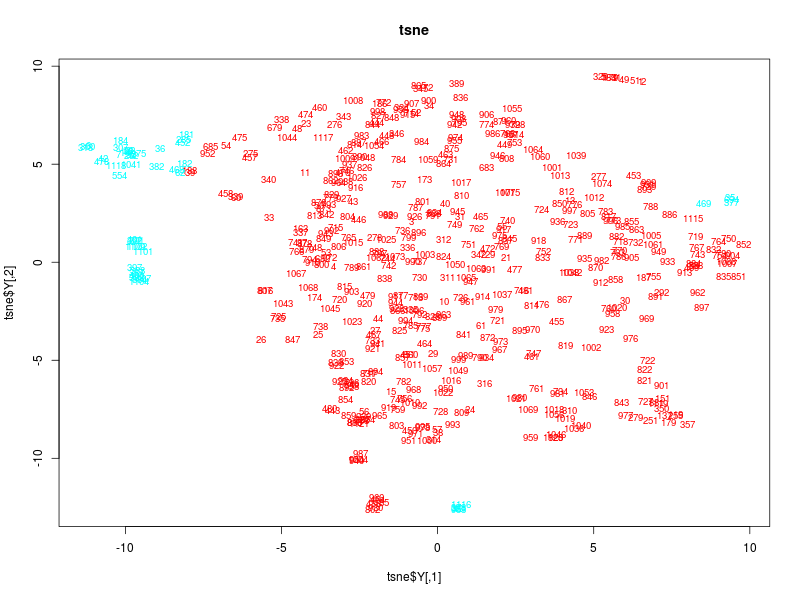}
\caption{Visualization with t-SNE algorithm for data sets clustering with an additional noise feature.}   
\label{fig:tsne_noise}
\end{figure} 

As can be seen from Figure~\ref{fig:tsne_noise}, t-SNE takes into account the random feature while constructing a visualization, but the border between clusters remains stable even with increasing levels of noise. Including noise results in clusters that are more scattered but still cleanly separated from each other. From this, we conclude that the number of clusters was chosen in a correct way and remains stable even in the presence of noise.

\subsubsection{Data set properties of the clusters} 
\label{subsubsec:propert_clust} 

Having clustered the data sets, the next stage of analysis is to study the properties of the data sets within these clusters. This can be done in numerous ways, but in this study, we focus on the data set characteristics discussed previously, in Table~\ref{tab:attributes}. We show the mean and median values of these characteristics in Table~\ref{t:mean_clusters} and Table \ref{t:median_clusters}, respectively. 
Considering both mean and median values is important, since in some cases, characteristics of our data sets have extreme values that can distort the mean while not being reflected by the median.

\begin{table}[!htb]
\centering
\caption{Mean values per cluster}
\label{t:mean_clusters}
\begin{tabular}{|p{5cm}|p{2cm}|p{2cm}|}
\hline

Name of parameter               & Cluster 1                         & Cluster 2                                    \\ \hline

NumAttributes                   & 30.457                            & 1625.26                                             \\ \hline

ClassEntropy                    & 0.885                             & 2.771                                             \\ \hline

DecisionStumpAUC                & 0.723                             & 0.73                                             \\ \hline

MaxNominalAttDistinctValues     & 36.88                             & 460.14                                             \\ \hline

MajorityClassSize               & 1790.292                          & 993.5                                             \\ \hline

\end{tabular}
\end{table}

\begin{table}[!htb] 
\centering
\caption{Median values per cluster}
\label{t:median_clusters}
\begin{tabular}{|p{5cm}|p{2cm}|p{2cm}|}
\hline

Name of parameter               & Cluster 1                         & Cluster 2                                    \\ \hline

NumAttributes                   & 11                                & 63.5                                             \\ \hline

ClassEntropy                    & 0.983                             & 2.799                                             \\ \hline

DecisionStumpAUC                & 0.719                             & 0.721                                              \\ \hline

MaxNominalAttDistinctValues     & -1                                & -1                                              \\ \hline

MajorityClassSize               & 203                               & 200                                              \\ \hline

\end{tabular}
\end{table} 

As shown in Table~\ref{t:mean_clusters}, Cluster~2 consists of large data sets with more than 1600 attributes, while the average length of the feature vectors in Cluster~1 is around 30. Since class entropy tends to be related to the number of features, it is not surprising that Cluster~2 also shows higher levels of class entropy.
Significant differences can be observed between the mean and median number of attributes (and class entropy).
The performance of a Decision Stump classifier for both clusters is close, but slightly higher for Cluster~2. Not only is the average number of features lower in Cluster~1, but its nominal features also have fewer values than the largest nominal features in another cluster. 
However, the median values for both clusters are once again quite close to each other. Finally, the majority class is almost two times larger in Cluster~2 than in Cluster~1. Based on the small difference in the median values seen in Table~\ref{t:median_clusters}, we conclude that this difference is caused by outliers.

\subsubsection{Selecting representative data sets for further analysis} 
\label{subsubsec:selec_dat_set} 

Next, we select ten representative data sets for further analysis from each of our two clusters using medoids. Taking into account the difference in size between the two clusters, we pick nine data sets from the larger Cluster~1 and one set from Cluster~2 (proportional to cluster size).
Medoids are central objects in a cluster and can be computed via Partitioning Around Medoids (PAM) analysis. 
PAM analysis \citep{kaufman1990partitioning} can help to identify the central elements in subclusters of our two clusters.
For Cluster~1, this results in the following choice of data sets:
`980\_optdigits', `844\_breastTumor', `751\_fri\_c4\_1000\_10', `831\_autoMpg', `1038\_gina\_agnostic', `457\_prnn\_cushings', `1119\_adult-census', `9\_autos', and `454\_analcatdata\_halloffame'. 
The single data set determined to be most representative of Cluster~2 is `20\_mfeat-pixel'. Details of these data sets can be found on the OpenML website, and the numbers represent OpenML IDs.

\subsection{Pareto Front analysis for algorithms}
\label{subsec:pareto_front}

Quantitative comparisons of classification procedures are typically based on Accuracy, FScore \citep{beitzel2006understanding} or AUC, and developers as well as users of such procedures typically focus on maximizing one of these measures.
Our frugality score is based on AUC and running time, and it is therefore natural and necessary to study the trade-off between the objectives of maximizing AUC and minimizing running time.
To this end, we investigated the respective Pareto fronts \citep{tomoiagua2013pareto}, where each point on front corresponds to a classification procedure whose performance in terms of AUC and running time is not dominated by any other classifier considered in our study.
In our plots of Pareto fronts, an ideal classifier would be represented by a point in the lower left corner of the plot, corresponding to maximum prediction accuracy (AUC) at minimum cost (running time).

\begin{figure}[t] 
\centering 
\includegraphics[scale=0.5]{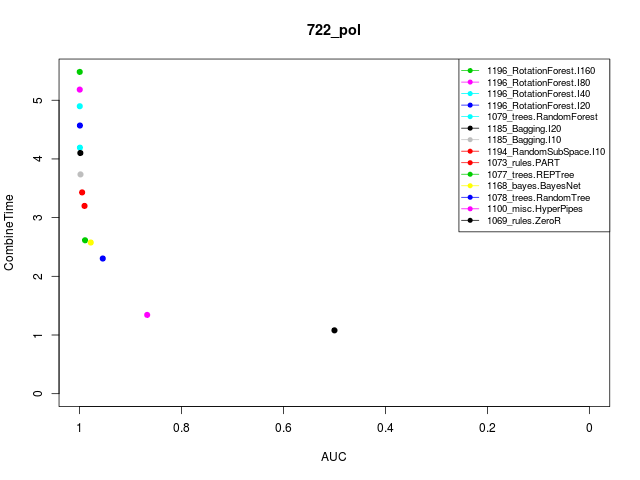}
\caption{Individual Pareto front for the 722\_pol data set. The Y-axis shows milliseconds on a $log_{10}$ scale}  
\label{fig:pareto_ind}
\end{figure}  

We constructed and studied the Pareto fronts for each data set individually. As an example, the Pareto front for the \verb|722_pol| data set is shown in Figure~\ref{fig:pareto_ind}. The best result for this data set in terms of AUC is achieved by the Rotation Forest algorithm with the number of iterations set to 160. However, it takes more than $10^5$ milliseconds to train and run the corresponding model. In contrast, the HyperPipes classifier completes the task almost instantly but with an AUC of about 0.87 (still significantly better than random guessing). The performance difference between these two algorithms is approximately 0.12 in terms of AUC and five orders of magnitude in terms of combined training and running time. All other algorithms show either lower computing time and lower AUC compared to Rotation Forest (I160) or better AUC and higher computing times than HyperPipes.\footnote{ZeroR is a base-line procedure that predicts the majority class and cannot be considered a meaningful classifier.}
Thus, based on given constraints on running time or requirements for AUC, the most suitable classifiers can be easily identified from the Pareto front.

\begin{figure}[t] 
\centering 
\includegraphics[scale=0.4]{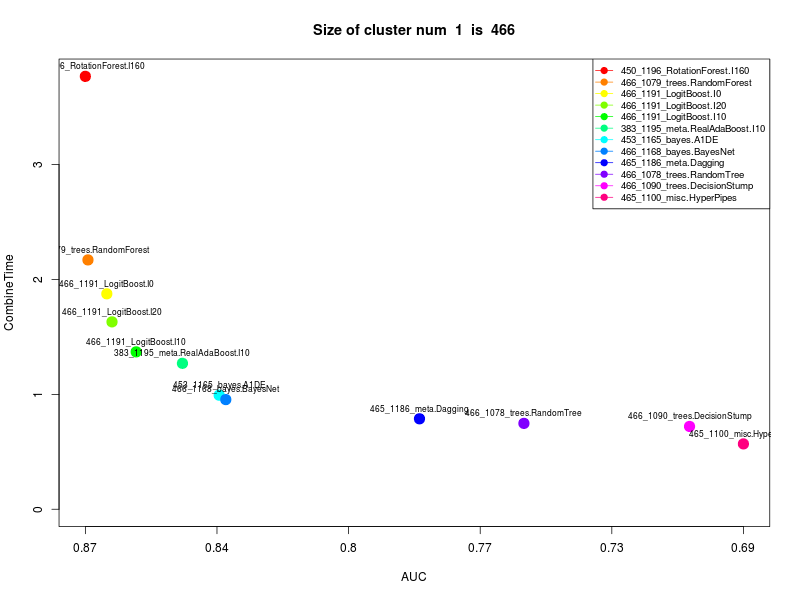} 
\caption{Pareto front for algorithms computer for Cluster 1, with 466 data sets and 15 algorithms on Pareto front.}   
\label{fig:pareto_cluster_first}
\end{figure}

\begin{figure}[t] 
\centering 
\includegraphics[scale=0.4]{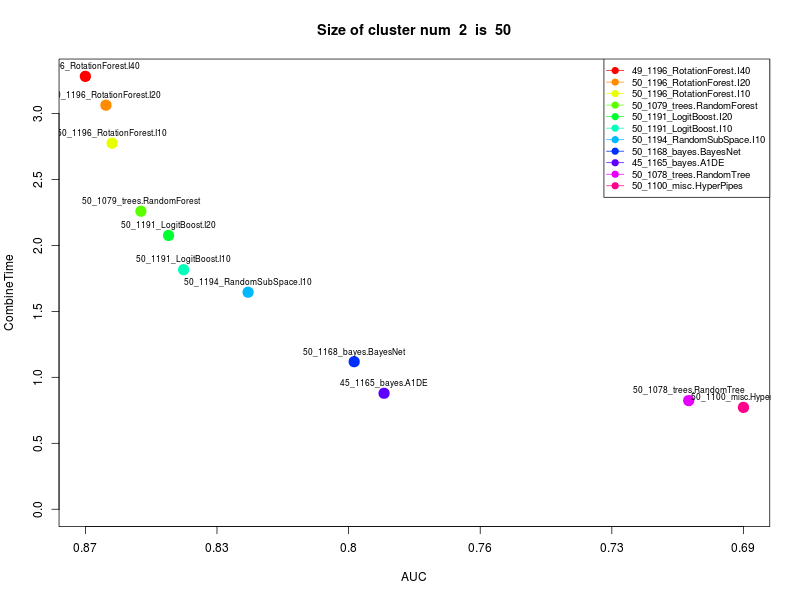} 
\caption{The second cluster with 50 data sets and Pareto front consisting of 14 algorithms.}   
\label{fig:pareto_cluster_second}
\end{figure}

Next, because the performances of our classifiers differ considerably between data sets, we clustered our  data sets (using kMeans with $k = 2$) as discussed in \ref{subsec:an_data_set} and averaged the CPU time and AUC per algorithm over each cluster. This gives us an ‘average’ picture of performance, for which we can again study the Pareto front. Averaged results presented in Figure \ref{fig:pareto_cluster_first} and \ref{fig:pareto_cluster_second}.  

From the Pareto front shown in Figure~\ref{fig:pareto_cluster_first}, we can see that Rotation Forest, Random Forest, and Boosted Stump classifiers are the best choices in the absence of strong constraints on running time. Otherwise, CPU time can be traded for AUC, and the A1DE algorithm may become an attractive choice. A1DE is a ensemble of 1-dependence classifiers, resulting in a Naive Bayes-like models with weaker feature independence assumptions \citep{Webb2005}.
If running time is severely limited, DecisionTree or HyperPipes classifiers become the methods of choice. Similar observations can be made for the second cluster, with some notable differences. As seen in Figure~\ref{fig:pareto_cluster_second}, a Rotation Forest with 40 iterations achieves the highest AUC, while for Cluster~1, a Rotation Forest with 160 iterations performed best in terms of AUC.
As shown in Section~\ref{subsubsec:propert_clust}, Cluster~2 contains data sets with more attributes. This could be a reason why Rotation Forest I160 is slow in this environment. While HyperPipes once again turns out to be fastest, Na\"{i}ve Bayes classifiers are now an attractive choice for intermediate trade-offs between the two performance objectives.

\subsection{Hierarchical clustering for algorithms} 
\label{subsec:hierar_clust_algorithms}
Next, we present the frugality scores for the algorithms in the form of heat maps, with parameter $w$ set to values of 0.1, 0.5, and 1, reflecting a low, moderate, or high cost of computational resources, respectively. 

\subsubsection{Transition to a low-dimensional feature space} 
\label{subsec:trans_low_dim}

\begin{figure}[t] 
\centering
\includegraphics[scale=0.09]{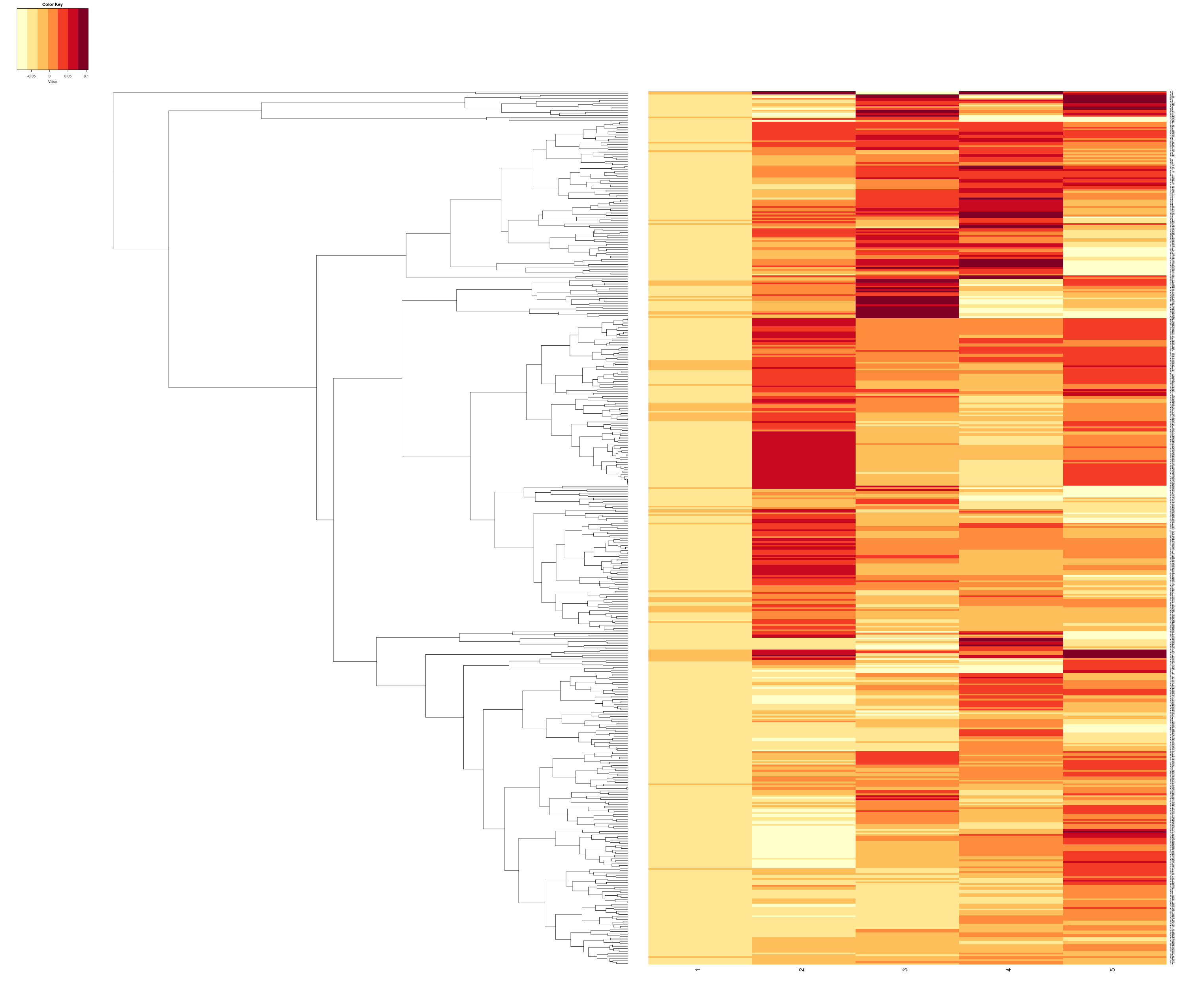}
\caption{Dendrogram for data sets in a low-dimensional space.}   
\label{fig:data_sets_svd} 
\end{figure}

As a first step towards producing heat maps, a hierarchical clustering of datasets is performed on the full table of frugality results. This creates a logical ordering of data sets, meaning that similar data sets are close to each other in the heat map. Because the original space created by the 103 algorithms is very high-dimensional, we first transform it to a lower-dimensional space by doing a singular value decomposition, and cancelling out the lowest singular values. The SVD transforms the original data set of frugality scores (with $w=0$) to a product of 3 matrices $U D V$, where $U$ represents data sets, $V^t $describes algorithms, and $D$ shows the importance of each (latent) dimension. The number of latent features can be chosen, and our experiments show that 5 latent features already explain about 91 percent of variance in our data. We therefore perform an SVD with 5 latent features to obtain a matrix $U$ with 516 rows and 5 columns. Next, we construct a hierarchical clustering over the data sets in $U$. The resulting dendrogram and 5-dimensional matrix of latent features are shown in Fig. \ref{fig:data_sets_svd}.

\subsubsection{Heat maps for the preserved order of rows and columns} 
\label{subsubsec:hierar_clust} 

We can now construct heatmaps in which the order of the datasets is directly derived from this dendrogram, to ensure that similar data sets are shown in spatial proximity to each other. The order of algorithms in our heat maps is also fixed and corresponds to the order in which they appear on the Pareto front built on all the datasets (combining Clusters 1 and 2): the leftmost algorithm has lowest AUC and fastest running time, while the rightmost algorithm has the highest AUC value and requires the largest amount of CPU time.

\setcounter{figure}{11}
\begin{figure}[p]
\subfloat[Frugality scores on all datasets for w=0.1]{\includegraphics[height=9.5cm]{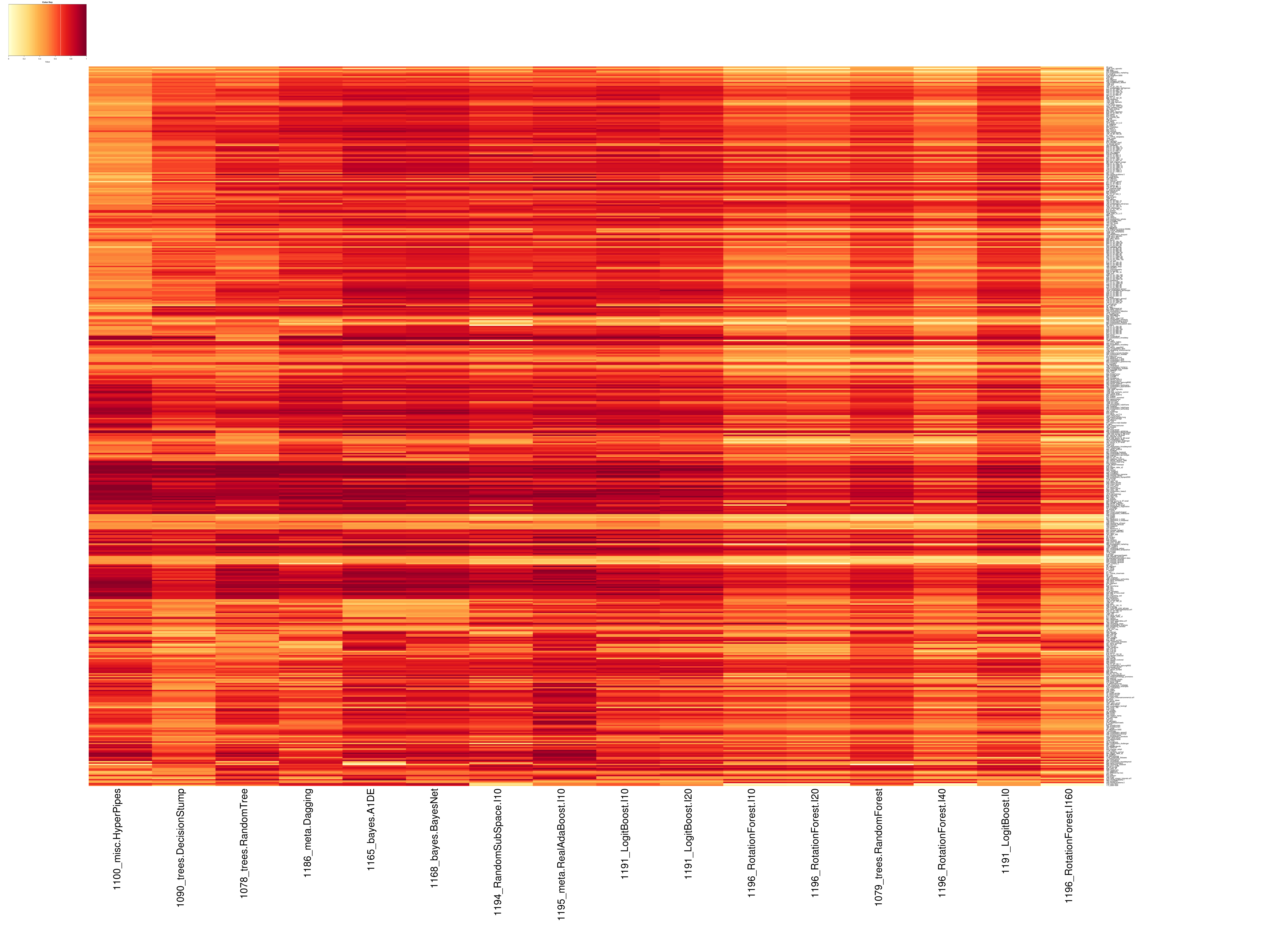} } \,
\subfloat[Frugality scores on all datasets for w=0.5]{\includegraphics[height=9.5cm]{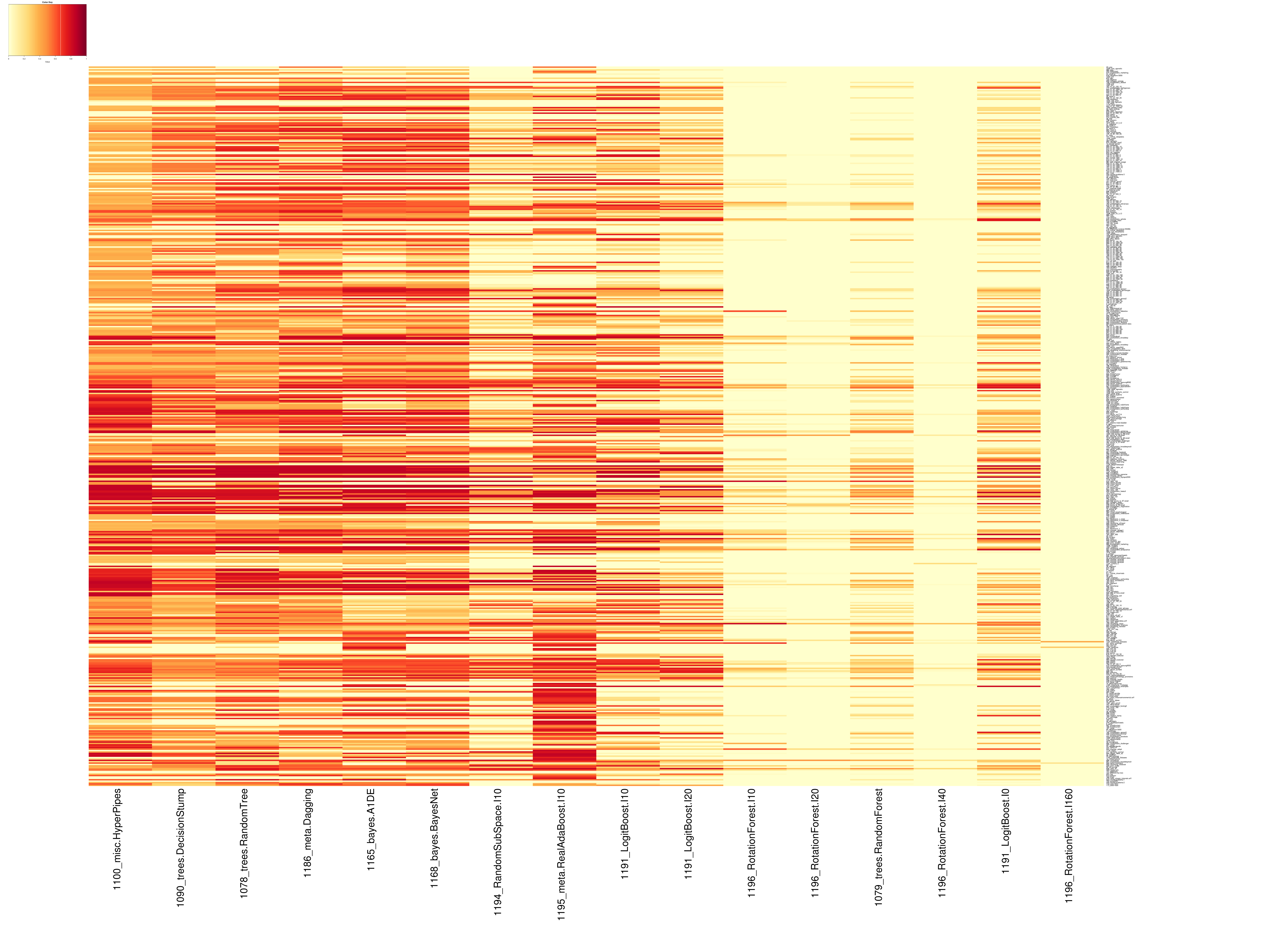} } \,
\end{figure}
\begin{figure}[t] 
\ContinuedFloat
\subfloat[Frugality scores on all datasets for w=1.0]{\includegraphics[height=9.5cm]{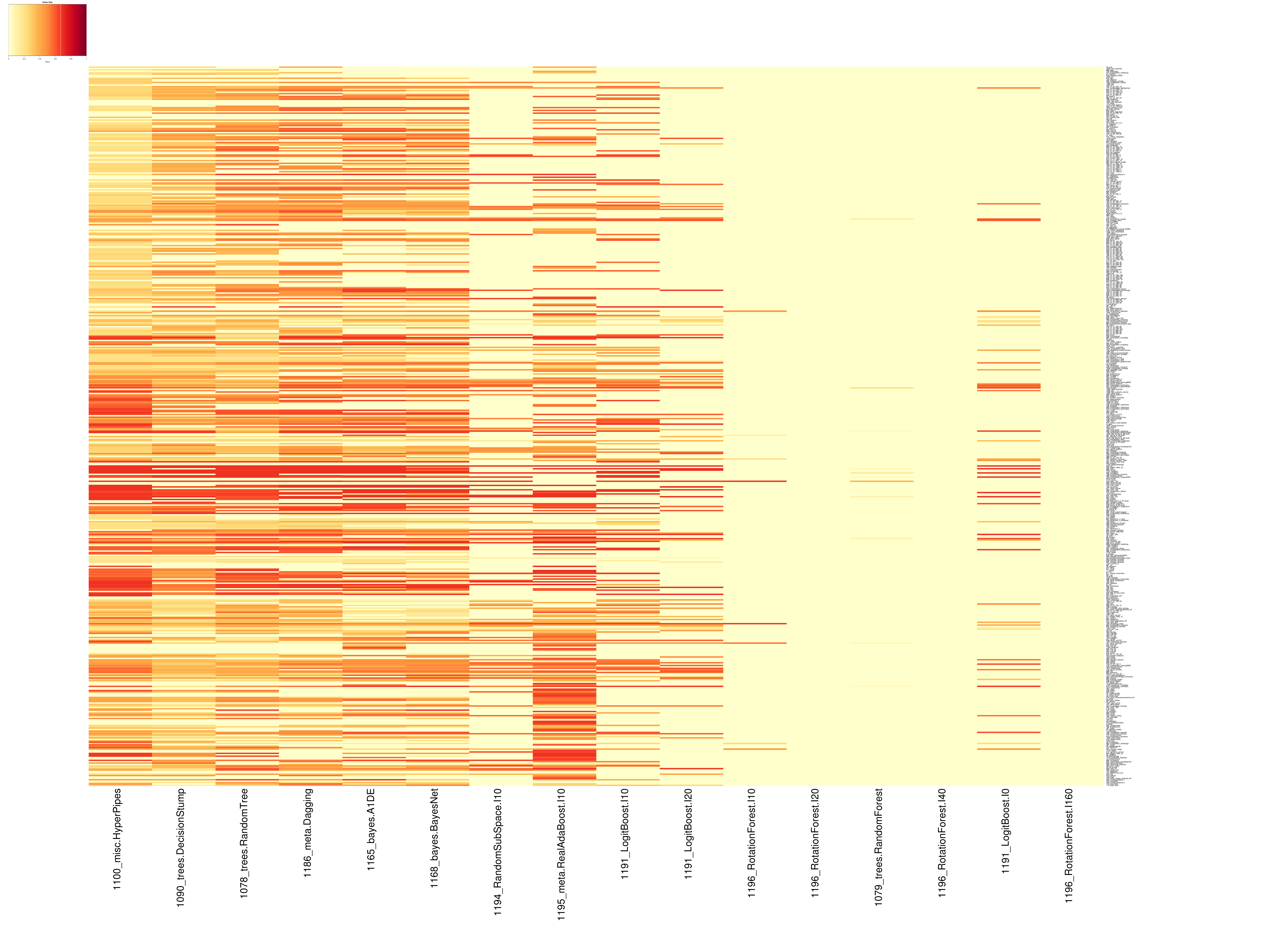} } \,
\caption{Hierarchical clustering results for increasing levels of frugality.}
\label{fig:allmaps}
\end{figure}

The frugality scores shown in the resulting heat maps, shown in Figure~\ref{fig:allmaps}, are coded such that high scores appear red and lower scores are represented by lighter colours. 
We show heat maps for different values of the frugality parameter $w$.
As can be seen in Figure~\ref{fig:allmaps}, as $w$ increases, frugality scores decrease. 
The extent to which this happens differs between algorithms, and some algorithms suffer more than others from an increase in the cost of computation. Compare, for instance, Random SubSpace with 10 iterations \emph{vs} Real AdaBoost with 10 iteration). In general, the high-performing algorithms on the right degrade fastest, although some algorithms, such as Logistic Boosting with 10 trees, survive quite well, at least on some datasets. Some of fast but less accurate algorithms on the left, such as `HyperPipes' and `RandomTree', maintain their scores quite well over a large number of datasets.

\subsection{Performance evaluation} 
\label{subsec:perf_evaluat}
Finally, we can study the frugality of our selected algorithms on the subset of representative datasets. We average the frugality scores over the 10 data sets identified in Section \ref{subsubsec:selec_dat_set}, and vary the frugality level $w$ from 0.0, meaning that there is no penalty for time at all, to 1.0, meaning that demand for frugality is very strong (i.e., algorithms should be able to work fast still showing decent performance).
 
\begin{figure}[!htb] 
\includegraphics[scale=0.45]{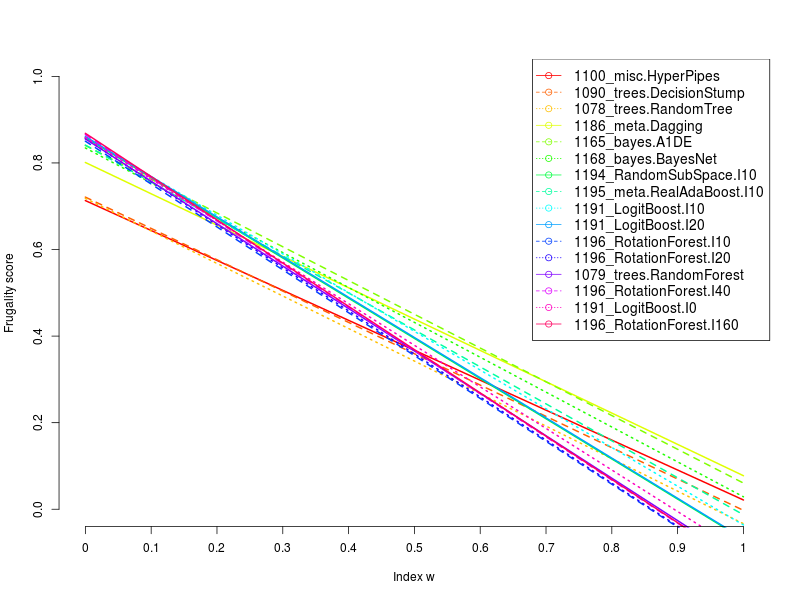}
\caption{Results for selected algorithms and data sets that are medoids for clusters. The numbers in the legend are the OpenML ID's.}  
\label{fig:lines}
\end{figure} 

The resulting frugality plots are shown in Figure \ref{fig:lines}. These can be used to find which algorithms should be considered when dealing with increasingly resource-scarce scenarios. A first lesson is that there are many similarly performing algorithms when CPU time is not an issue. However, many algorithms quickly fall down the ranking if frugality comes into play, especially large ensembles. In general, the ranking of frugal algorithms seems to slowly flip, with the best algorithms becoming the worst and vice versa. Looking at the results for LogitBoost (Logistic Boosting with, in this case, Decision Stumps), we can observe that the version with 20 iterations is quickly overtaken by the version with 10 iterations. 

The most frugal algorithm for $w=0$ is a large Rotation Forest with 160 trees, but from $w=0.1$ it is replaced by Logistic Boosting (first with 20 and then with 10 iterations). In turn, these are overtaken by A1DE (a Naive Bayes-like learner) around $w=0.2$, which stays the most frugal learning algorithm right up to $w=0.7$, when it is replace by Dagging. The latter is an ensemble learner that splits the data into stratified chunks, and gives each to a base classifier (here a Decision Stump). Another algorithm that performs quite well (in second or third place for most of the range) is BayesNet, a Bayesian Network learner. At extreme levels of frugality, much higher than $w=1$, HyperPipes ultimately takes the crown, but by then frugality has dropped below 0. 

Overall, a general lesson would be that A1DE, boosted stumps with few iterations, and Bayesian Networks are good choices for frugal learning.

\section{Future work}
\label{sec:future}
In future work, we aim to evaluate the most promising algorithms (in terms of frugality) on a smartwatch, to solve real-world tasks such as Human Activity Recognition (HAR), and study the performance and battery consumption of these algorithms. Smartwatches have plenty of sensors such as an accelerometer, gyroscope, gravity sensor, linear acceleration sensor, rotation vector, step detector, air pressure sensor, magnetometer, and heart rate monitor. The last one is especially important since most of the HAR research performed with consumer wearable devices is based on this data. The set of activities that we want to classify based on this data are: walking, walking downstairs, walking upstairs, sitting, standing, and laying down. These activities are also used for activity recognition on smartphones \citep{anguita2013public}. 

\begin{figure}[t] 
\centering 
\includegraphics[scale=0.4]{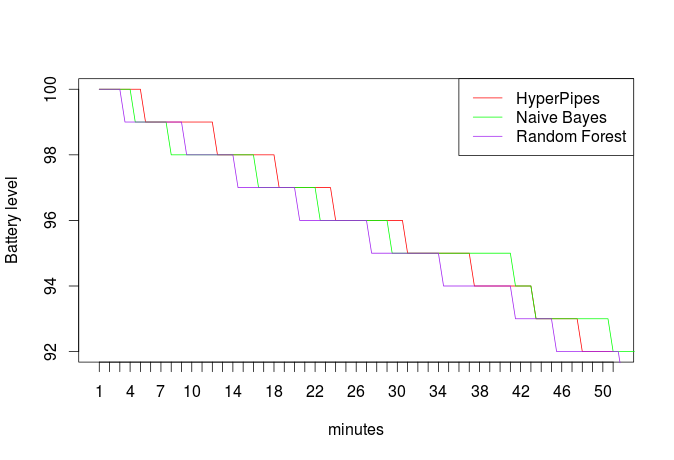} 
\caption{Battery drain for HyperPipes, Na\"{i}ve Bayes, and Random Forest}   
\label{fig:battery}
\end{figure}

Based on earlier analysis, interesting candidates seem to be Na\"{i}ve Bayes (based on the A1DE results), HyperPipes and Random Forest. The latter is not very frugal but is included as a baseline. We have already ported these algorithms on an Android Wear device and have started to collect data. Figure \ref{fig:battery}, for instance, shows the change in battery level when performing activity recognition with HyperPipes, Na\"{i}ve Bayes, and Random Forest algorithms. While we run the HAR task, we sample battery charge twice per minute, and keep the application running for 50 minutes. We repeat this experiment for every algorithm, fully recharging the device before every run. The watch was placed on the wrist, and the application was run in ambient mode\footnote{http://developer.android.com/training/wearables/apps/always-on.html} all the time.     

Random Forests lead to the fastest battery drain, corroborating our earlier findings. HyperPipes has slower battery drain, while NaiveBayes' results seem to fluctuate. Future work will pursue larger, more definite experiments.

There are several other avenues for future work. Most importantly, we aim to study \textit{frugal stream mining}, studying the frugality of stream mining algorithms that are able to handle concept drift. We also aim to extend our analysis to study the impact of algorithm hyperparameters, as well as the impact of preprocessing techniques such as feature selection.

\section{Conclusions}
\label{sec:conclusion}
With the rapid rise of wearable devices and the Internet of Things, many new machine learning applications will pervade our everyday lives. These applications may not always permit streaming large amounts of data over a network, and even if they do, privacy issues may come into play. Indeed, 
while wearable sensors will enable us to improve our quality of life, we are not all comfortable sharing significant amounts of our location, movement and biometric data on a regular basis. Hence, we can expect an increasing demand for machine learning to be performed on wearable and embedded devices themselves. 

We presented a new framework for the analysis of machine learning algorithms in terms of their \textit{frugality}, i.e., of how proficient they are at delivering accurate predictions when working with (possibly severely) limited resources. We introduced a novel evaluation measure, the frugality score, which trades off predictive accuracy for resource consumption and can be adjusted to the resources available to a learning algorithm, depending, for instance, on whether it is to be run on a smartphone or a smartwatch.

In an extensive empirical analysis, evaluating 103 learning algorithms on 517 classification data sets, we discovered that our collection of data sets can be divided into two clearly separate clusters, on which the algorithms perform quite differently. Using a Pareto Front analysis, we found which algorithms are the most interesting candidates for frugal learning, and we visualized their frugality scores on all data sets for increasingly resource-scarce scenario's. Finally, we introduced \textit{frugality curves}, which allow to quickly discover how different algorithms perform relative to each other, and find that algorithms that perform best without resource limitations are often the least efficient, and vice versa. In particular, we found that while large ensembles perform best when given ample resources, they are quickly overtaken by Naive Bayes-like algorithms and boosted decision stumps with few iterations.

In future work, we aim to extend this analysis by porting these learning algorithms to smartwatches and measuring their actual performance on wearable devices. Preliminary results do corroborate our earlier findings. 

The concept of frugality offers an interesting and useful new way to analyse learning algorithm performance. By studying their performance through a `frugal lens', we can discover many new interesting properties and leverage this knowledge to field machine learning algorithms in pervasive low-power scenarios, and stimulate research into novel, more frugal machine learning algorithms.

\bibliographystyle{plainnat}
\bibliography{sources} 

\clearpage 

\begin{appendices} 

\section{} 
\label{appendix:algorithms_studied} 
\begin{longtable}
{|p{7cm}|p{5cm}|}  
\caption*{ Algorithms and used parameters, blank parameters value in case default values} 
\label{t:attributes} 
\endhead 
\hline 
ID and a name of Algorithm on OpenML & Used parameters  \\ \hline  
1068\_trees.J48 &  \\ \hline  
1070\_rules.Ridor &  \\ \hline  
1071\_rules.OneR &  \\ \hline  
1072\_rules.Prism &  \\ \hline  
1073\_rules.PART &  \\ \hline  
1074\_rules.OLM &  \\ \hline  
1075\_rules.NNge &  \\ \hline  
1076\_trees.SimpleCart &  \\ \hline  
1077\_trees.REPTree &  \\ \hline  
1078\_trees.RandomTree &  \\ \hline  
1079\_trees.RandomForest &  \\ \hline  
1080\_trees.NBTree &  \\ \hline  
1082\_trees.LMT &  \\ \hline  
1084\_trees.J48graft &  \\ \hline  
1086\_trees.Id3 &  \\ \hline  
1087\_trees.HoeffdingTree &  \\ \hline  
1088\_trees.FT &  \\ \hline  
1089\_trees.ExtraTree &  \\ \hline  
1090\_trees.DecisionStump &  \\ \hline  
1091\_trees.BFTree &  \\ \hline  
1094\_rules.JRip &  \\ \hline  
1096\_rules.DecisionTable &  \\ \hline  
1098\_rules.ConjunctiveRule &  \\ \hline  
1099\_misc.CHIRP &  \\ \hline  
1100\_misc.HyperPipes &  \\ \hline  
1101\_misc.InputMappedClassifier &  \\ \hline  
1102\_misc.OSDL &  \\ \hline  
1103\_misc.VFI &  \\ \hline  
1104\_lazy.IB1 &  \\ \hline  
1105\_lazy.IBk & K1, K3, K5 \\ \hline  
1106\_lazy.KStar &  \\ \hline  
1107\_lazy.LBR &  \\ \hline  
1108\_lazy.LWL &  \\ \hline  
1109\_functions.GaussianProcesses &  \\ \hline  
1111\_functions.KernelLogisticRegression &  \\ \hline  
1112\_LibLINEAR & E0.001, E0.01 \\ \hline  
1114\_functions.Logistic &   \\ \hline  
1117\_functions.RBFClassifier &  \\ \hline  
1120\_functions.SGD & \\ \hline  
1154\_bayes.NaiveBayes &  \\ \hline  
1160\_meta.AttributeSelectedClassifier &  \\ \hline  
1165\_bayes.A1DE &  \\ \hline  
1166\_bayes.A2DE & \\ \hline  
1168\_bayes.BayesNet &  \\ \hline  
1172\_functions.LibSVM & \\ \hline  
1173\_bayes.HNB &  \\ \hline  
1174\_functions.KernelLogisticRegression &  \\ \hline  
1177\_functions.SimpleLogistic &  \\ \hline  
1178\_functions.SMO &  \\ \hline  
1179\_functions.SMO & \\ \hline  
1180\_functions.SPegasos &  \\ \hline  
1181\_functions.Winnow &  \\ \hline  
1182\_AdaBoostM1 & I10, I20, I40, I80, I160  \\ \hline  
1183\_meta.AttributeSelectedClassifier & \\ \hline  
1185\_Bagging & I10, I20, I40, I80, I160   \\ \hline  
1186\_meta.Dagging & \\ \hline  
1187\_meta.Decorate & \\ \hline  
1188\_meta.END &  \\ \hline  
1190\_meta.Grading & \\ \hline  
1191\_LogitBoost & I0, I10, I20, I80, I160   \\ \hline   
1192\_MultiBoostAB & I10, I20, I40, I80, I160  \\ \hline   
1193\_meta.RacedIncrementalLogitBoost & \\ \hline  
1194\_RandomSubSpace & I10, I20, I40, I80, I160  \\ \hline  
1195\_meta.RealAdaBoost & I10, I20, I40, I80, I160   \\ \hline  
1196\_RotationForest & I10, I20, I40, I80, I160   \\ \hline  
1197\_rules.FURIA &  \\ \hline  
1199\_trees.ADTree &  \\ \hline  
1200\_trees.LADTree &  \\ \hline  
1244\_meta.FilteredClassifier &  \\ \hline 
\end{longtable} 

\clearpage 

\section{} 
\label{appendix:missing_values}  
  
\begin{table}[h]
\caption*{Names and the amount of unsuccessful runs per algorithm} 
\label{t:unattributes}
\begin{tabular}{|p{8cm}|p{3cm}|}
\hline
Name of algorithm           & Missing values \\ \hline    

1109\_functions.GaussianProcesses        & 516 \\ \hline  
1072\_rules.Prism                        & 491 \\ \hline 
1086\_trees.Id3                          & 489 \\ \hline 
1102\_misc.OSDL                          & 486 \\ \hline 
1173\_bayes.HNB                          & 486 \\ \hline 
1181\_functions.Winnow                   & 480 \\ \hline 
1107\_lazy.LBR                           & 461 \\ \hline 
1089\_trees.ExtraTree                    & 249 \\ \hline 
1111\_functions.KernelLogisticRegression & 121 \\ \hline 
1174\_functions.KernelLogisticRegression & 121 \\ \hline 
1180\_functions.SPegasos	             & 111 \\ \hline  
1195\_meta.RealAdaBoost.I160             & 111 \\ \hline 
1120\_functions.SGD                      & 110 \\ \hline 
1195\_meta.RealAdaBoost.I10              & 110 \\ \hline 
1195\_meta.RealAdaBoost.I20              & 110 \\ \hline 
1195\_meta.RealAdaBoost.I40              & 110 \\ \hline 
1195\_meta.RealAdaBoost.I80              & 110 \\ \hline 
1199\_trees.ADTree                       & 110 \\ \hline 
1166\_bayes.A2DE                         & 80 \\ \hline 
1099\_misc.CHIRP                         & 60 \\ \hline 
1075\_rules.NNge                         & 30 \\ \hline 
1196\_RotationForest.I160                & 22 \\ \hline 
1187\_meta.Decorate                      & 20 \\ \hline 
1196\_RotationForest.I80                 & 20 \\ \hline 
1165\_bayes.A1DE                         & 18 \\ \hline 
1114\_functions.Logistic                 & 14 \\ \hline 
1160\_meta.AttributeSelectedClassifier   & 11 \\ \hline 
\end{tabular}
\end{table} 

\end{appendices}

\end{document}